# Diffusion-based Radiotherapy Dose Prediction Guided by Inter-slice Aware Structure Encoding

Zhenghao Feng, Lu Wen, Jianghong Xiao, Yuanyuan Xu, Xi Wu, Jiliu Zhou, *Senior Member, IEEE,* Xingchen Peng, Yan Wang*, *Member, IEEE*

*Abstract*—Deep learning (DL) has successfully automated dose distribution prediction for radiotherapy planning, increasing both efficiency and quality. However, existing methods commonly utilize $L_1$ or $L_2$ loss to calculate the posterior average, thus heavily suffering from the over-smoothing problem. To address this, we propose a diffusion model-based method, named DiffDose, to automatically predict radiotherapy dose distribution for cancer patients. Specifically, our DiffDose model contains a forward process and a reverse process. In the forward process, DiffDose gradually adds small noise to dose distribution maps via multiple steps until converting them to pure Gaussian noise, and a noise predictor is simultaneously trained to estimate the noise added in each step. In the reverse process, DiffDose iteratively removes the noise from a pure Gaussian noise leveraging the well-trained noise predictor and finally outputs the predicted dose distribution maps. Concretely, to provide the model with essential structure information, we design a structure encoder to learn the anatomical knowledge from patients' anatomy images, guiding the noise predictor to generate dose distribution maps that are aware of personalized structures. Considering the latent continuity and similarity among sliced anatomy images, an inter-slice interaction transformer ($I^2T$) block is embedded in the structure encoder to capture such long-range dependency. Extensive experiments on an in-house dataset involving 130 rectum cancer cases validate the superiority of our method.

This work is supported by National Natural Science Foundation of China (NSFC 62371325, 62071314), Sichuan Science and Technology Program 2023YFG0025, 2023YFG0101, and 2023 Science and Technology Project of Sichuan Health Commission 23LCYJ002.

This work involved human subjects in its research. Approval of all ethical and experimental procedures and protocols was granted by the Biomedical Research Ethics Committee of the West China Hospital of Sichuan University (Chengdu, China).

Zhenghao Feng and Lu Wen contribute equally to this work. Yan Wang is the corresponding author.

Zhenghao Feng, Lu Wen, Yuanyuan Xu, and Yan Wang are with the School of Computer Science, Sichuan University, China. (e-mail: fzh_scu@163.com; wenlu0416@163.com; YuanyuanXuSCU@outlook.com; wangyanscu@hotmail.com).

Jianghong Xiao is with the Department of Radiation Oncology, Cancer Center, West China Hospital, Sichuan University, China. (e-mail: xiaojh86@foxmail.com).

Xi Wu and Jiliu Zhou are with the School of Computer Science, Chengdu University of Information Technology, China. (e-mail: wuxi@cuit.edu.cn; zhoujiliu@cuit.edu.cn).

Xingchen Peng is with the Department of Biotherapy, Cancer Center, West China Hospital, Sichuan University, China. (e-mail: pxx2014@scu.edu.cn).

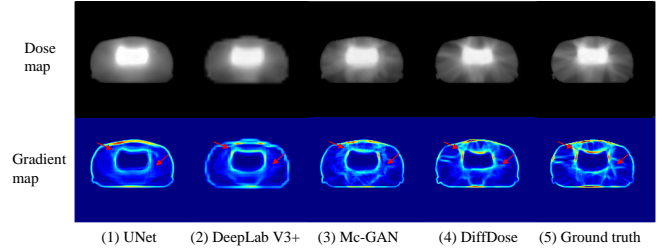

**Fig. 1.** Instances from a rectum cancer patient. The first row displays the dose maps predicted by UNet, DeepLab V3+, Mc-GAN, and the proposed DiffDose while the second row gives their corresponding gradient maps in Jet color style.

## I. INTRODUCTION

RADIOTHERAPY, standing as the indispensable clinical treatment for malignant tumors, has recently received significant advancements. A high-quality radiotherapy plan achieves the therapeutic effectiveness by delivering adequate radiation dose to the planning target volume (PTV) while minimizing radiation exposure to organs at risk (OARs). However, dosimetrists are strictly required to manually adjust these plans, making it a tedious and laborious trial-and-error process [1]. Moreover, variations of expertise among dosimetrists may lead to inconsistency in plan quality [2]. Therefore, it is essential to explore an automatic method for predicting dose distributions for cancer patients, which can expedite the radiotherapy process by offering dosimetrists a better starting point for treatment planning.

As emerging solutions, deep learning (DL)-based methods have recently made remarkable strides in automatic treatment planning by directly predicting the dose distribution maps [3-10, 25]. In this vein, Nguyen *et al.* [3] first used a modified 2D UNet to obtain an automatic prediction of the dose maps for prostate cancer patients. Utilizing UNet as the backbone, several works introduced various innovations for accuracy enhancements [5]-[6]. Besides, Song *et al.* [7] exploited DeepLab V3+ for predicting the dose distribution. Despite their impressive performance, the DL-based methods confront the over-smoothing problem. Concretely, they commonly use the content losses (i.e., $L_1$ and $L_2$ loss) for model optimization which computes a posterior mean of the joint distribution between the ground truth and predictions [12]. Thus, these models yield over-smoothed predicted dose maps to a greater or lesser extent, lacking crucial high-frequency details [13]. In Fig. 1, we present predicted dose maps from various deep models alongside their corresponding gradient maps generated by a simple Sobel operator to indicate the radiation patterns in the dose maps. As seen, compared to ground truth (i.e., Fig. 1(5)), predictions (1) and (2) appear evidently blurred with less high-frequency information, particularly for ray shapes. Such high-frequency details generated by ray penetration disclose the ray directions and dose attenuation for eliminating cancer cells while safeguarding OARs during radiotherapy. As for preserving high-frequency information, generative adversarial network (GAN) becomes a powerful option that uses adversarial training to incorporate an adversarial loss with the commonly used content losses, acquiring predicted results with higher quality [8]. Nevertheless, inheriting the common problem of GAN, such GAN-based dose prediction models are more difficult to train owing to the gradient disappearance or gradient explosion issues [14]. Besides, they usually introduce artifacts and unnatural textures as shown in Fig. 1(3). Therefore, exploring a method to automatically produce accurate predictions with sufficient high-frequency details is crucial for dose prediction.

Currently, the diffusion model [15]-[16] has shown its powerful capability for modeling complex image distributions in multiple vision tasks [17], [18]. Different from other DL-based methods, the diffusion model has two distinct advantages. First, without imposing additional assumptions about the distribution of target data, diffusion model evades the averaging effect and alleviates the issue of over-smoothing [17]. Second, compared to GANs, the continuous diffusion process enhances the convergence stability of the diffusion model, eschewing the gradient disappearance or gradient explosion [19]. However, the potential of diffusion model in radiotherapy dose prediction remains mostly undiscovered.

In this paper, we present a novel diffusion model-based method, namely DiffDose, to automatically predict the clinically acceptable dose distribution from the Computed Tomography (CT) image and segmentation masks of the PTV and OARs. DiffDose comprises a forward process and a reverse process. Specifically, in the forward process, the model progressively adds predefined noise to complexly distributed dose maps for multiple steps and finally converts them into standard Gaussian distribution, employing a Markov chain. Meanwhile, a noise predictor is trained to estimate the noise added in the corresponding step. In the reverse process, the model gradually removes the noise starting from pure Gaussian noise iteratively leveraging the well-trained noise predictor until the accurate prediction of dose map is ultimately generated. To effectively provide the noise predictor with crucial structure information, we devise a transformer-based structure encoder to learn the anatomical knowledge from the CT image, PTV and OARs masks, which indicates the structure and relative position of organs and tumor. By including such abundant structure information, the noise predictor focuses on the critical areas of PTV and OARs, thus delivering more appropriate dose to them and producing more accurate dose distribution maps. Besides, as with most existing dose prediction methods, the input images (i.e., CT images, segmentation mask of PTV and OARs) are sliced from 3D volumes, leading to a lack of inter-slice information brought by the inherent continuity and implicit similarity among slices. However, such information among neighboring slices is critical for generating an accurate dose distribution map, especially for certain radiology slices, e.g., apex and base slices. Therefore, we borrow the strong capability of the transformer in extracting global dependency and design an inter-slice interaction transformer ($I^2T$) block to learn such inter-slice information. By embedding the $I^2T$ blocks into the structure encoder, the inter-slice knowledge can be effectively introduced into the network. During the model optimization, traditional DL-based methods evaluate the discrepancy between the prediction and ground truth which treats each pixel prediction equally, thus neglecting the higher importance of some critical organs (i.e., PTV and OARs) to the radiotherapy efficacy. In this regard, we design an adaptive weighted loss to enforce the model to focus more on the PTV and OARs using more fine-grained dose constraints, thus improving the prediction quality.

The contributions of our work can be summarized as four-fold: (1) We present DiffDose, a diffusion-based model for dose prediction in radiotherapy to tackle the prevalent issue of over-smoothing existing in current DL-based radiotherapy dose prediction methods. (2) We design a structure encoder to thoroughly extract the anatomical knowledge from the CT images and organ segmentation masks, guiding the noise predictor in the diffusion model to form more accurate dose distribution maps. (3) We introduce an innovative inter-slice interaction transformer ($I^2T$) block in the structure encoder to exploit the latent continuity and similarity among slices, thus enhancing the predictions for apex and base slices. (4) Considering the higher importance of the PTV and OARs regions to the clinical efficacy, we present an adaptive weighted loss to exert ample attention to both PTV and OARs for an overall performance enhancement. Extensive experiments on a clinical rectum cancer dataset have verified the superiority of our approach.

Notably, the preliminary version of this work was published at the *26th International Conference on Medical Image Computing and Computer-Assisted Intervention (MICCAI 2023)* [10]. We extend the conference paper in the following four aspects: (1) Introduction: we elaborate on the research background to comprehensively introduce the motivations driving this work; (2) Related works: we add a dedicated section to review relevant literature; (3) Methodology: we further introduce an $I^2T$ block to model the inter-slice continuity and similarity and design an adaptive weighted loss to take the higher importance of PTV and OARs regions into account. Besides, a more efficient structure encoder is devised to extract the anatomical information. (4) Experiments: we add two more state-of-the-art methods, i.e., U-ResNet-D [20] and Mc-GAN [21], to validate the effectiveness of our methods and devise more persuasive ablation studies to demonstrate the contribution of each component. More clinic metrics for comprehensive evaluation are also incorporated.

## II. RELATED WORKS

### A. Deep Learning in Radiotherapy Dose Prediction

Deep learning (DL) has revolutionized the automatic prediction method of dose distribution for its outstanding ability of feature extraction. As one of the most popular encoder-decoder architectures, UNet has been widely applied to dose prediction tasks. Liu *et al.* [20] constructed a 3D UNet-like network (U-ResNet-D) to obtain an accurate prediction of the dose distribution maps of head-and-neck (H&N) cancer. Liu *et al.* [5] introduced cascade 3D UNet (C3D) for dose prediction. Wang *et al.* [6] exploited a progressive refinement UNet, named PRUNet, to gradually rectify the predictions in different resolutions. Additionally, Song *et al.* [7] applied another classical encoder-decoder network, i.e., DeepLab V3+, to accelerate the radiotherapy planning of rectum cancer. The DeepLabV3+ could extract more global context information from multiple scales with atrous spatial pyramid pooling (ASPP), further improving the prediction performance. Treating UNet as the generator, several GAN-based methods have also been proposed [8], [11], [21]. Kearney *et al.* [11] designed an attention-gated GAN to diminish the network redundancy through only concentrating on relevant anatomy. More recently, Zhan *et al.* [21] devised a multi-organ constraint GAN (Mc-GAN) to force the network to further consider the dose requirements of PTV and OARs and obtain more clinically acceptable dose distribution.

The existing DL-based methods always depend on content loss (sometimes incorporating an adversarial loss) to constrain the similarity between the ground truth and prediction, leading to unavoidable over-smoothing problems. Differently, we exploit the diffusion model to fulfill the dose prediction task. Without relying on additional assumptions about the distribution of target data, our method can effectively alleviate the over-smoothing issue and form high-quality predictions.

### B. Diffusion Model in Medical Image Processing

The Diffusion model was first presented by Sohl-Dickstein *et al.* [19] for deep unsupervised learning and then brought notable breakthroughs in several medical image processing tasks, i.e., modality translation [22], segmentation [23], and denoising [24]. For instance, Lyu *et al.* [22] took advantage of the diffusion model to fulfill the modality translation of image modalities, i.e., from Magnetic Resonance Imaging (MRI) to CT, gaining better synthesis performance than GAN-based methods. Kim *et al.* [23] proposed a diffusion adversarial representation learning (DARL) model which utilizes a diffusion module to learn the background signal and a generation module to output the segmentation predictions of vessel masks. Xiang *et al.* [24] imposed a self-supervised statistic-based denoising method into the diffusion model to form clean images through the conditional generation process in an unsupervised manner. Plenty of works have verified the effectiveness of the diffusion model in modeling complex distribution and generating more realistic outputs. Inspired by this, we also explore the practicality of employing a diffusion model for dose prediction task, which is expected to maintain more high-frequency information in the dose maps, thus enhancing the overall performance.

## III. METHODOLOGY

An overview of the proposed DiffDose model is displayed in Fig. 2 (A). Following Denoising Diffusion Probabilistic Models (DDPM) [16], our DiffDose also comprises two Markov chain processes, i.e., a forward process and a reverse process. An image pair of a patient is defined as $\{x, y\}$, where $x \in \mathbb{R}^{(2+o) \times H \times W}$ denotes the structure images, "2" denotes the channels of CT image and the segmentation mask of the PTV, while $o$ indicates the total number of the OAR masks. $y \in \mathbb{R}^{1 \times H \times W}$ is the corresponding dose distribution map for $x$. In the forward process, noisy images $\{y_0, y_1, \ldots, y_T\}$ ($y_0 = y$) are generated by progressively adding a small amount of noise to $y_0$ across $T$ steps ($T$ is theoretically large enough) where the noise intensity is increased step by step, until a pure Gaussian noise $y_T$ is gained. Meanwhile, a noise predictor $f$ is designed and trained to estimate the noise added to $y_{t-1}$. To gain the anatomical knowledge, a structure encoder $g$ is constructed to extract the vital feature representation $x_e$ from the structure images. Inter-slice interaction transformer (I$^2$T) block is designed and injected into the structure encoder to compensate for the lost inter-slice knowledge when slicing 3D volume into 2D images. Incorporating the anatomical feature $x_e$ with $y_t$ and current noise intensity $\gamma_t$ to form the input, the noise predictor is trained to concentrate more on the critical anatomical knowledge highly related to the dose prediction. Subsequently, in the reverse process, the model gradually infers the dose distribution map through iteratively denoising from $y_T$ with the well-trained noise predictor.

### A. Diffusion Model

The detailed framework of DiffDose is devised following the classical DDPM [16] which includes a forward process and a reverse process, in which DiffDose progressively converts the Gaussian noise into dose distribution maps with complex data distribution.

**Forward Process:** In this process, the DiffDose model utilizes the Markov chain to gradually add noise to the initial input $y_0 \sim q(y_0)$ until the final disturbed image $y_T$ becomes pure Gaussian noise which is denoted as $y_T \sim \mathcal{N}(y_T \mid 0, I)$, which is formulated as:

$$q(y_{1:T} \mid y_0) \coloneqq \Pi_{t=1}^{T} q(y_t \mid y_{t-1}), \tag{1}$$

$$q(y_t \mid y_{t-1}) \coloneqq \mathcal{N}(y_t; \sqrt{\alpha_t} y_{t-1}, (1 - \alpha_t)I), \tag{2}$$

where $\alpha_t \in (0,1)$ is the unlearnable standard deviation of the noise added to $y_{t-1}$.

Herein, the $\alpha_t$ ($t = 1, \ldots, T$) increases during the forward process, and we define the noise intensity as $\gamma_t = \prod_{i=1}^{t} \alpha_i$. Then, we can directly achieve the distribution of $y_t$ at any step $t$ from $y_0$ using the following single-step calculation:

$$q(y_t \mid y_0) = \mathcal{N}(y_t; \sqrt{\gamma_t} y_0, (1 - \gamma_t)I), \tag{3}$$

where the disturbed image $y_t$ is sampled using:

$$y_t = \sqrt{\gamma_t} y_0 + \sqrt{1 - \gamma_t} \varepsilon_t, \tag{4}$$

in which $\varepsilon_t \sim \mathcal{N}(0, I)$ is random noise sampled from normal

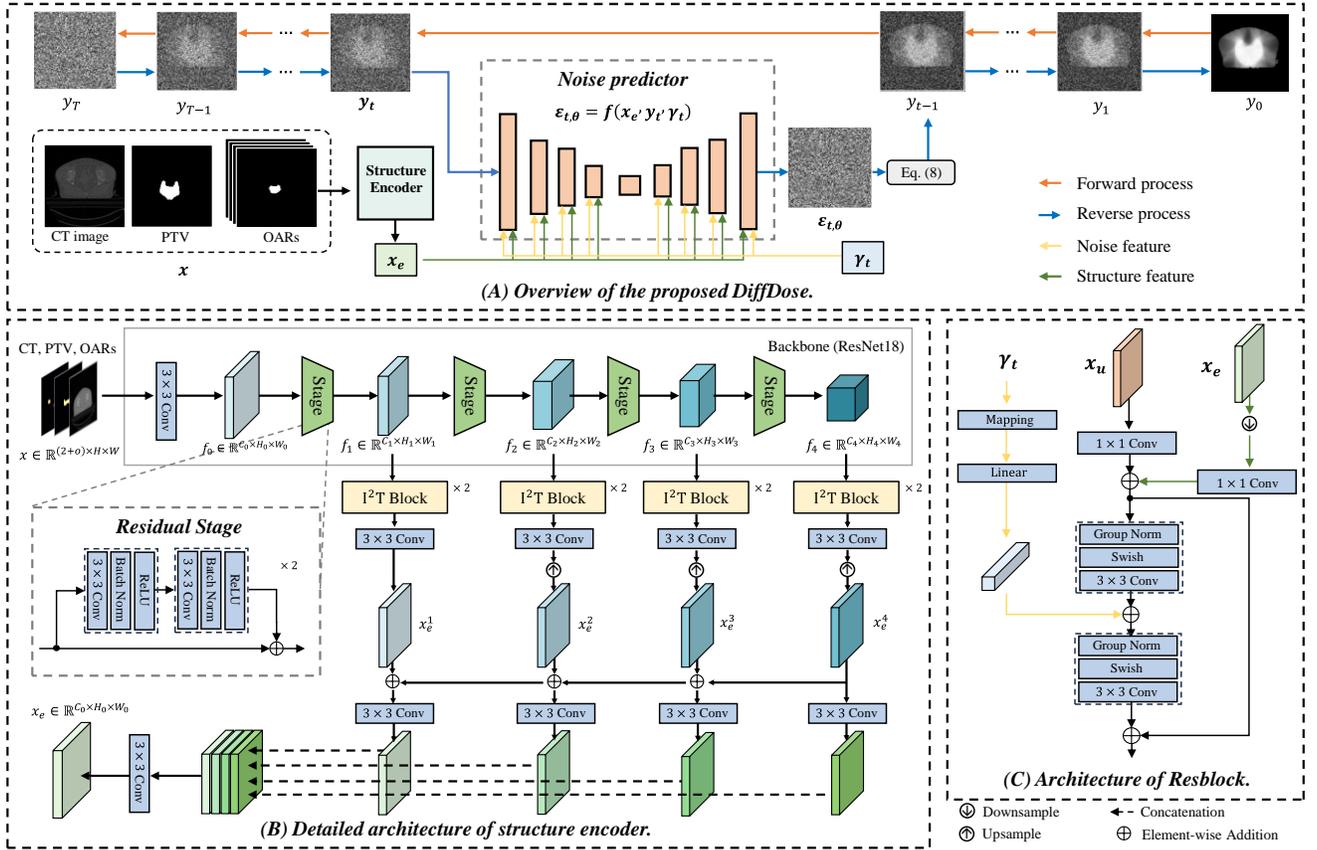

**Fig. 2.** Overview of the proposed DiffDose is shown in (A) which involves a forward process for model training (i.e., converting the dose map into Gaussian noise) and a reverse process for testing (i.e., predicting the dose map from Gaussian noise). DiffDose is equipped with a transformer-based structure encoder (as shown in (B)) for gaining anatomical information. (C) gives a detailed architecture of Resblock in the noise predictor.

Gaussian distribution.

**Reverse Process:** Consistent with the forward process, the reverse process also uses the Markov chain to progressively transform the latent variable distribution $p_\theta(y_T)$ back to the distribution of dose maps $p_\theta(y_0)$ parameterized by $\theta$. The reverse process is a denoising transformation conditioned on structure images $x$ and starts from a pure Gaussian noise $y_T \sim \mathcal{N}(y_T \mid 0, I)$, which is formulated as below:

$$p_\theta(y_{0:T} \mid y_t, x) = p(y_T) \prod_{t=1}^{T} p_\theta(y_{t-1} \mid y_t, x), \quad (5)$$
$$p_\theta(y_{t-1} \mid y_t, x) = \mathcal{N}(y_{t-1}; \mu_\theta(x, y_t, \gamma_t), \sigma_t^2 I), \quad (6)$$

where $\sigma_t$ is an unlearnable standard deviation and the mean $\mu_\theta(x, y_t, t)$ depends on the noise predictor (Described in Section III.C). Drawing from [16], we parameterize the mean $\mu_\theta$ as below:

$$\mu_\theta(x, y_t, \gamma_t) = \frac{1}{\sqrt{\alpha_t}} \left( y_t - \frac{1-\alpha_t}{\sqrt{1-\gamma_t}} \varepsilon_{t,\theta} \right), \quad (7)$$

where $\varepsilon_{t,\theta}$ is a function aimed at predicting $\varepsilon_t$ based on the input of $x$, $y_t$, and $\gamma_t$. Therefore, the reverse inference for two adjacent steps can be formulated as:

$$y_{t-1} \leftarrow \frac{1}{\sqrt{\alpha_t}} \left( y_t - \frac{1-\alpha_t}{\sqrt{1-\gamma_t}} \varepsilon_{t,\theta} \right) + \sqrt{1-\alpha_t} z_t, \quad (8)$$

where $z_t \sim \mathcal{N}(0, I)$ denotes randomly sampled noise from normal Gaussian distribution.

*B. Structure Encoder*

**Architecture of Structure Encoder:** Vanilla diffusion model struggles to retain vital structure information and often yields suboptimal results when directly generating dose distribution maps from noise with plain condition mechanisms such as directly concatenating $x$ and $y_t$ across channels for the noise predictor inputting. To tackle this, we devise a structure encoder $g$ that learns the anatomical information from CT image and segmentation masks of PTV and OARs to provide the noise predictor with structure knowledge, thus enhancing the accuracy of generated dose maps. The architecture of the structure encoder is displayed in Fig. 1(B) which employs the ResNet-18 [26] as the backbone. We substitute the first $7 \times 7$ convolution layer and max pool layer in ResNet-18 with a $3 \times 3$ convolution layer to maintain the scale. Besides the first convolutional layer, the ResNet-18 includes four identical residual stages and each has four $3 \times 3$ convolution blocks where residual connections are reserved between every two convolution blocks for avoiding gradient vanishment during the training procedure. ResNet-18 takes structure image $x \in \mathbb{R}^{(2+o) \times H \times W}$ as input, encompassing the CT image along with segmentation masks of PTV and OARs, and produces multi-level feature representation $f_i \in \mathbb{R}^{C_i \times H_i \times W_i}$, $i \in \{1, 2, 3, 4\}$ in the $i$-th stages. To capture the essential inter-slice information, $f_i$ is subsequently fed into two I²T blocks to capture such

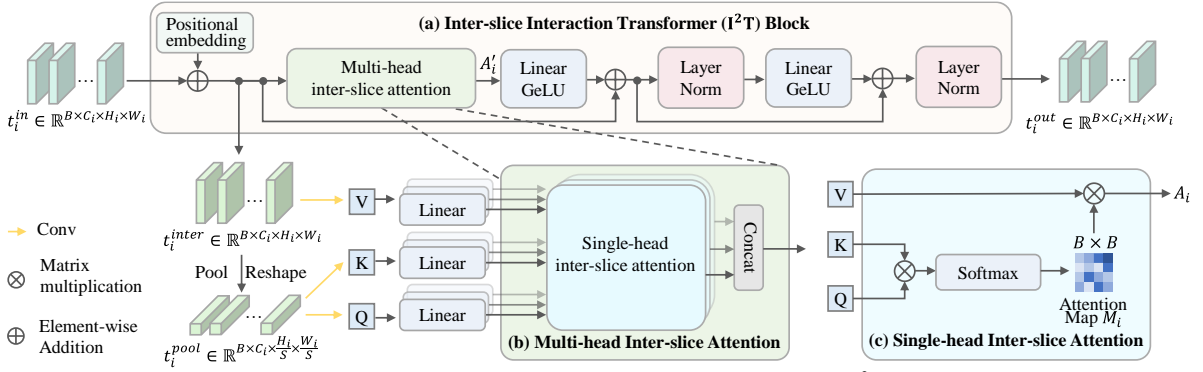
**Fig. 3.** Detailed structure of the inter-slice interaction transformer (I$^2$T) block.

long-term dependency. Instead of directly using features from the very last layer of the backbone, we aggregate the features with different spatial resolutions learned from all intermediate residual stages. Specifically, the feature of the deeper stage is incorporated into that of the shallower stage for effective information utilization. Then, the output feature maps of different stages are up-sampled to the same dimension, i.e., $x_e^i \in \mathbb{R}^{C_0 \times H_0 \times W_0}$ ($C_0 = 64$). Finally, a concatenation is used to aggregate the multi-level information in $x_e^i$ and a convolution operation is used to produce the final structure feature $x_e \in \mathbb{R}^{C_0 \times H_0 \times W_0}$ which is then fed into the noise predictor.

**Inter-slice Interaction Transformer Block:** Considering the continuity and similarity among slices are critical for predicting the accurate dose distribution, we introduce an inter-slice interaction transformer (I$^2$T) block to model such long-range dependency. The structure of the I$^2$T block is depicted in Fig. 3(A), with a multi-head inter-slice attention (MHIA) sub-block in each. As seen, treating the 4D tensor $t_i^{in} \in \mathbb{R}^{B \times C_i \times H_i \times W_i}$ with a stack of $B$ feature maps (i.e., $f_i^b \in \mathbb{R}^{C_i \times H_i \times W_i}$, b$\in [1, B]$) in the minibatch, we first incorporate it with positional embeddings and form an intermediate feature $t_i^{inter} \in \mathbb{R}^{B \times C_i \times H_i \times W_i}$, following the philosophy of transformer [27]. However, different from the original transformer, we design the MHIA to replace the traditional multi-head self-attention to further capture the inter-slice information. Representing the output of MHIA as $A_i'$, the process of I$^2$T block to form its output $t_i^{out} \in \mathbb{R}^{B \times C_i \times H_i \times W_i}$ can be stated as:

$$t_i^{out} = LN\big(GeLU(t_i^{temp} W_2 + b_2) + t_i^{temp}\big), \quad (9)$$
$$t_i^{temp} = LN\big(GeLU(A_i' W_1 + b_1) + t_i^{inter}\big), \quad (10)$$

where $W_1$ and $W_2$ denote the linear projection, $b_1$ and $b_2$ are bias terms, $LN$ represents layer normalization, and $GeLU$ is the Gaussian error linear unit. After processing by the I$^2$T block, we restore $t_i^{out}$ to $B$ feature maps equipped with rich inter-slice interaction information and feed them into the rest part of the structure encoder.

Considering MHIA is the main component of the I$^2$T block, we give its detailed architecture in Fig. 3(B) which contains $H$ ($H = 4$ in this work) single-head inter-slice attention (SHIA) sub-blocks as shown in Fig. 3(C). We first reshape $t_i^{inter}$ using average pooling with a kernel size of $S$ ($S = 4$ in this work) and gain a condensed stack of feature maps $t_i^{pool} \in \mathbb{R}^{B \times C_i \times \frac{H_i}{S} \times \frac{W_i}{S}}$. Then, each SHIA takes three different projections from $t_i^{iner}$ and $t_i^{pool}$, i.e., queries ($Q$), keys ($K$), and values ($V$), as inputs, which can be described as below:

$$Q = t_i^{pool} W_Q, \ K = t_i^{pool} W_K, V = t_i^{temp} W_V, \quad (11)$$

where $W_Q, W_K$, and $W_V \in \mathbb{R}^{C_i \times C_i}$ denote the learnable linear projection matrices of $Q$, $K$, and $V$. Notably, $Q$ and $K$ are linear projections of $t_i^{pool}$ while $V$ is derived from $t_i^{inter}$. The output of SHIA (i.e., $A_i \in \mathbb{R}^{B \times C_i \times H_i \times W_i}$) is calculated as:

$$A_i = SHIA(Q, K, V) = MV,$$
$$M = softmax\big(\frac{QK^T}{\sqrt{C_i H_i W_i / S^2}}\big), \quad (12)$$

where $M \in \mathbb{R}^{B \times B}$ is an attention map and each element $M[j, k]$ in it evaluates the similarity between the query of the $j$-th slice and the key of the $k$-th slice. So, while processing a slice, $M$ represents how much attention the model should pay to other slices, explicitly modeling the interaction among slices.

After obtaining the outputs of $H$ SHIA heads, i.e., $\{A_i^h\}_{h=1}^H$, the MHIA block concatenates them in the $C_i$ dimension, thus gaining the final output, i.e., $A_i'$, as below:

$$A_i' = MHIA(Q, K, V) = Concat(A_i^1, A_i^2, \ldots, A_i^H) W_3, \quad (13)$$

where $W_3$ is a learnable parameter matrix.

*C. Noise Predictor*

The noise predictor, denoted as $f(x_e, y_t, \gamma_t)$, is responsible for predicting the noise added to the dose distribution map $y_t$ conditioned on the anatomical feature $x_e$ from the structure images $x$ and noise intensity $\gamma_t$ at each step $t$. Following Ho *et al.* [26], we utilize a four-level UNet [28] and replace the original convolution blocks with residual ones to construct the noise predictor. Specifically, jointly input with noisy image $y_t$, the extracted feature $x_e$ and noise intensity $\gamma_t$, the traditional encoding-decoding procedure is used to extract the deep feature from the noisy image $y_t$. The structure information in $x$ is enforced to be transferred into the noise predictor. As displayed in Fig. 2(C), for each residual block inside the UNet, the structure feature $x_e$ is down-sampled to the corresponding dimension with convolution operations and then fused with the output of the residual block, i.e., $x_u$, through a simple element-wise addition. The skip connections are retained for effective integrating and reusing of high- and low-level information. Finally, the noise predictor produces the final output, i.e., the noise $\varepsilon_{t,\theta} = f(x_e, y_t, \gamma_t)$ in step $t$.

## D. Objective Function

The objective function of the DiffDose is with the goal of optimizing the noise predictor $f$ and structure encoder $g$, ensuring the predicted noise $\varepsilon_{t,\theta}$ during the reverse process approximates the added noise $\varepsilon_t$ in the forward process. To this end, the traditional diffusion model employs a simple $L_1$ loss to measure the pixel-level discrepancy. However, such loss with equal punishment to all pixels cannot reflect the higher importance of PTV and OAR regions to the final radiotherapy efficacy. Concretely, the clinical dose requirements, i.e., sufficient dose to PTV while as low dose to OARs as possible, indicate that accurate dose distribution within PTV and OARs is important for obtaining the radiotherapy efficacy, requiring more concentration than other regions. Therefore, we present an adaptive weighted loss as

$$\min_{\theta} \mathbb{E}_{(x,y)} \mathbb{E}_{\varepsilon,\gamma} \sum_{(j,k) \in \varepsilon_{t,\theta}} \lambda^{j,k} \|\varepsilon_{t,\theta} - \varepsilon_t\|, \varepsilon_t \sim \mathcal{N}(0,I), \quad (14)$$

$$\varepsilon_{t,\theta} = f\left(g(x), \underbrace{\sqrt{\gamma}y_0 + \sqrt{1-\gamma}\varepsilon_t}_{y_t}, \gamma_t\right), \quad (15)$$

with a weighted term

$$\lambda^{j,k} = \begin{cases} 1, & \text{if } (j,k) \in Mask_{Background} \\ \lambda_1^{j,k}, & \text{else if } (j,k) \in Mask_{PTV} \cup Mask_{OARs}, \\ \lambda_2^{j,k}, & \text{otherwise} \end{cases} \quad (16)$$

where $(j,k)$ represents the pixel coordinate and $Mask_{PTV}$, $Mask_{OARs}$, and $Mask_{Background}$ denote the region of PTV, OARs, and background, respectively. By assigning higher values to $\lambda_1^{j,k}$ and $\lambda_2^{j,k}$, the model concentrates more on the PTV and OARs, thus predicting more accurate dose maps.

## IV. EXPERIMENTS AND RESULTS

### A. Dataset Description and Evaluation Metrics

**Dataset:** We validate our method with an in-house dataset with 130 rectum cancer patients (87 male and 43 female) who received volumetric modulated arc therapy (VMAT) treatment at West China Hospital. Specifically, each case includes the CT images, PTV mask, OAR masks, and clinically planned dose distribution. All CT images are scanned with 3mm thickness with patients in the supine position. The PTV and OARs segmentation masks are contoured by experienced oncologists and finally approved through the international consensus guidelines [29]. The four OARs of rectum cancer are the bladder (BLD), FemoralHeadR (FHR), FemoralHeadL (FHL), and small intestine (ST). The PTV is imposed on a prescribed dose of 50.40Gy/28 fractions. All the plans are executed using the Raystation v4.7 TPS with Elekta Versa HD linear accelerators.

For data partition, we randomly choose 98 patients for model training, 10 patients for validation, and the remaining 22 patients for testing. Before the training process, all the 3D volumes with a resolution of 3mm×3mm×3mm are sliced into continuous 2D images with a size of 160×160 along the axial direction. In this way, we finally gain 16346, 1529, and 3491 2D slices for training, validation, and testing, respectively.

**Evaluation Metrics:** We extensively evaluate our proposed model using various clinical and statistical metrics including $D_x$, $V_x$, conformity index ($CI$) [30], heterogeneity index ($HI$) [31] and dose score [32]. $D_x$ represents the minimum absorbed dose that covers $x\%$ volume of PTV or OARs, while $V_x$ denotes the percentage volume that receives a dose level of at least $x$. $D_{mean}$ denotes the mean dose absorbed.

Concretely, $HI$ is defined as:

$$HI = \frac{D_2 - D_{98}}{D_{50}}. \quad (17)$$

$CI$ can be expressed as:

$$CI = \frac{(TV \cap PIV)^2}{TV \times PIV}, \quad (18)$$

where $TV$ denotes the target volume of PTV while $PIV$ represents the prescription isodose volume.

According to the clinical standard, we use $D_{98}$, $D_{50}$, $D_2$, $D_{mean}$, $CI$, and $HI$ for dose evaluation of PTV. For OARs, $D_2$, $D_{mean}$, and $V_{40}$ are selected. For a more direct quantification, we calculate the average prediction errors (APEs) of all metrics as below:

$$|\Delta M| = \frac{1}{n} \sum_{i=1}^{n} |M_{gt}^i - M_{pre}^i|, \quad (19)$$

where $M_{gt}^i$ and $M_{pre}^i$ denote the ground truth and predicted metrics of the $i^{th}$ patient, respectively, and $n$ is the number of patients in the testing dataset. For a general comparison, the dose score [32] is also utilized to directly evaluate the mean dose error between the clinically planned ground truth $D_{gt}^i$ and predicted dose distribution $D_{pre}^i$ within regions of interest (ROIs) over all patients in the testing dataset as follows:

$$\text{dose score} = \frac{1}{n} \sum_{i=1}^{n} |D_{gt}^i - D_{pre}^i| * Mask_{ROI}, \quad (20)$$

where $Mask_{ROI}$ represents the binary segmentation mask of body, PTV, or OARs.

For a more intuitive view, we incorporate the dose volume histogram (DVH) [33] as an additional metric for assessing prediction performance. Enhanced accuracy of dose distribution is indicated when the DVH curves of the predictions closely align with the ground truth.

TABLE I
QUANTITATIVE RESULTS OF SOTA METHODS IN TERMS OF DOSE SCORE↓(GY), SHOWN IN THE MEAN (STANDARD DEVIATION) FORM. THE BEST RESULT OF EACH INDEX IS MARKED IN **BOLD** WHILE THE SECOND-BEST ONE IS UNDERLINED. * MEANS OUR METHOD IS SIGNIFICANTLY BETTER THAN COMPARED METHOD WITH P < 0.05 VIA PAIRED T-TEST.

| ROI | UNet | U-ResNet-D | DeepLabV3+ | C3D | PRUNet | Mc-GAN | DiffDP | Proposed |
|---|---|---|---|---|---|---|---|---|
| Body | 2.184(0.338)* | 2.141(0.332)* | 1.837(0.332)* | 1.655(0.314) | 1.795(0.383)* | 1.749(0.280)* | 2.148(0.589)* | **1.615(0.257)** |
| PTV | 2.298(1.346)* | 2.511(1.505)* | 1.423(1.318)* | 1.436(1.170)* | 1.425(1.402)* | 1.562(1.004)* | **1.191(1.162)** | 1.196(0.368) |
| ST | 5.014(1.610)* | 3.811(1.012)* | 3.176(1.186)* | 3.326(1.406)* | 3.359(1.464)* | 3.044(0.943)* | 3.836(2.064)* | **2.361(0.476)** |
| FHL | 5.378(2.632)* | 4.622(2.052)* | 3.550(1.426)* | 3.436(1.307)* | 3.660(1.598)* | 3.422(1.428)* | 4.054(2.080)* | **2.370(0.693)** |
| FHR | 5.210(2.479)* | 4.048(1.626)* | 3.199(1.237)* | 3.490(1.687)* | 3.160(1.097)* | 3.290(1.444)* | 3.594(1.393)* | **2.225(0.528)** |
| BLD | 6.597(2.100)* | 3.886(1.174)* | 3.516(1.415)* | 3.524(1.158)* | 3.639(1.506)* | 3.813(1.634)* | 3.656(1.283)* | **2.582(0.547)** |

TABLE II
QUANTITATIVE RESULTS OF SOTA METHODS IN TERMS OF $|\Delta D_x|$, $|\Delta HI|$ AND $|\Delta CI|$, SHOWN IN THE MEAN (STANDARD DEVIATION) FORM. THE BEST RESULT OF EACH INDEX IS MARKED IN **BOLD** WHILE THE SECOND-BEST ONE IS UNDERLINED.

| ROI | Metrics | UNet | U-ResNet-D | DeepLabV3+ | C3D | PRUNet | Mc-GAN | DiffDP | Proposed |
|---|---|---|---|---|---|---|---|---|---|
| PTV | $|\Delta D_{98}|\downarrow$(Gy) | 3.751(3.064)* | 3.505(3.248)* | 2.305(3.324)* | 2.554(2.759)* | 2.265(3.391)* | 2.202(3.095) | **1.902(3.093)** | 2.056(2.537) |
|  | $|\Delta D_{50}|\downarrow$(Gy) | 2.221(1.468)* | 2.556(1.684)* | 0.975(1.593)* | 0.801(1.388)* | 0.914(1.638)* | 1.273(1.215)* | 0.776(1.353)* | **0.479(0.359)** |
|  | $|\Delta D_2|\downarrow$(Gy) | 1.043(0.383)* | 1.043(0.383)* | 0.547(0.462) | 0.561(0.476) | 0.754(0.422)* | 1.004(0.458)* | **0.486(0.350)** | 0.562(0.356) |
|  | $|\Delta D_{mean}|\downarrow$(Gy) | 2.077(1.352)* | 2.416(1.504)* | 0.968(1.450)* | 0.856(1.185)* | 0.959(1.465)* | 1.268(1.049)* | 0.780(1.180)* | **0.493(0.389)** |
|  | $|\Delta HI|\downarrow$ | 6.14E-2 (6.17E-2)* | 5.79E-2 (5.99E-2)* | 4.49E-2 (6.08E-2)* | 5.43E-2 (5.17E-2)* | 4.20E-2 (6.24E-2)* | 3.98E-2 (6.47E-2) | **3.63E-2 (6.34E-2)** | 4.50E-2 (4.20E-2) |
|  | $|\Delta CI|\downarrow$ | 0.131 (0.136)* | 8.68E-2 (0.127)* | 4.50E-2 (0.142)* | 6.35E-2 (0.144)* | 6.51E-2 (0.136)* | 8.27E-2 (0.150)* | 6.83E-2 (0.131)* | **3.18E-2 (4.17E-2)** |
| ST | $|\Delta D_2|\downarrow$(Gy) | 2.390(1.852)* | 2.610(1.764)* | 0.983(1.660)* | 0.882(1.260)* | 1.231(1.819)* | 1.168(1.457)* | 1.139(1.495)* | **0.379(0.329)** |
|  | $|\Delta D_{mean}|\downarrow$(Gy) | 4.116(2.277)* | 2.250(1.634)* | 1.708(1.576)* | 2.324(1.808)* | 2.071(1.768)* | 1.635(1.267)* | 2.786(1.954)* | **0.743(0.664)** |
| FHL | $|\Delta D_2|\downarrow$(Gy) | 6.476(4.202)* | 6.525(3.698)* | 4.046(3.171)* | 4.000(3.224)* | 3.872(3.656)* | 3.573(2.923)* | 3.714(3.031)* | **2.191(1.785)** |
|  | $|\Delta D_{mean}|\downarrow$(Gy) | 4.962(2.883)* | 3.558(2.660)* | 2.432(1.851)* | 2.272(1.833)* | 2.747(2.095)* | 2.498(1.724)* | 3.072(2.683)* | **1.115(1.072)** |
| FHR | $|\Delta D_2|\downarrow$(Gy) | 4.999(4.796)* | 4.309(3.604)* | 4.211(2.734)* | 4.988(4.090)* | 4.313(2.415)* | 4.135(2.824)* | 3.995(2.593)* | **1.882(1.860)** |
|  | $|\Delta D_{mean}|\downarrow$(Gy) | 4.880(2.648)* | 3.109(2.130)* | 2.032(1.483)* | 2.351(2.196)* | 1.986(1.450)* | 2.128(1.836)* | 2.431(1.940)* | **0.892(0.698)** |
| BLD | $|\Delta D_2|\downarrow$(Gy) | 1.958(1.973)* | 1.928(1.872)* | 1.146(1.657)* | 1.089(1.548)* | 1.054(1.943)* | 1.295(1.315)* | 0.903(1.310)* | **0.549(0.392)** |
|  | $|\Delta D_{mean}|\downarrow$(Gy) | 6.369(2.249)* | 2.694(1.582)* | 2.464(1.846)* | 2.576(1.419)* | 2.528(1.987)* | 2.717(1.987)* | 1.852(1.584)* | **0.994(0.819)** |

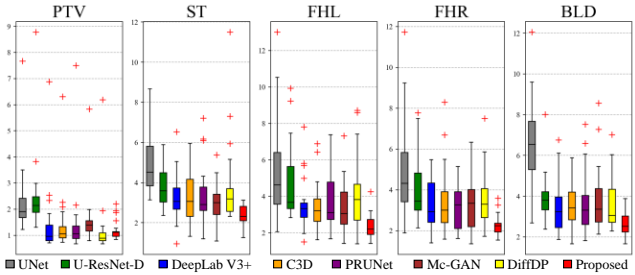

**Fig. 4.** Dose errors of SOTA methods on PTV and OARs.

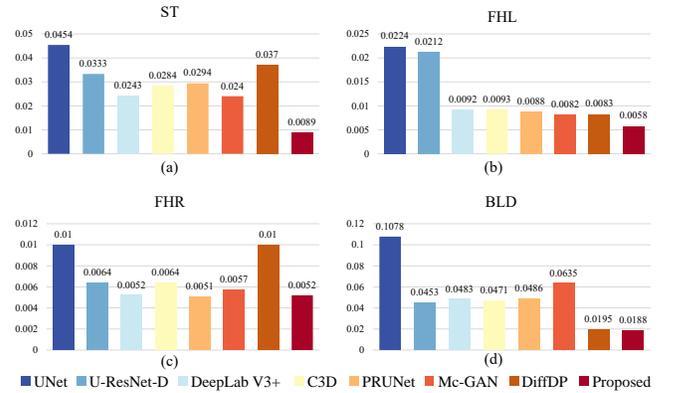

**Fig. 5.** Quantitative results of SOTA methods on OARs (i.e., ST, FHL, FHR, and BLD) in terms of $|\Delta V_{40}|$.

## B. Implementation Details

We implement our proposed network with the PyTorch framework, and all experiments are conducted on a single NVIDIA GeForce RTX 3090 GPU with 24GB memory. The training stage is performed with a batch size of 16 using the Adaptive Moment Estimation (Adam) optimizer for 500 epochs, where the learning rate is set to 1e-4. $\lambda_1^{j,k}$ and $\lambda_2^{j,k}$ in (16) are set to 4 and 2, respectively. The parameter $T$ is set to 1000 for training. Additionally, the noise intensity $\gamma_t$ is set following the cosine schedule [34]. Notably, the continuous slices are partitioned from the 3D volume and input into the model. For the inference (reverse) process, to make a trade-off between the time cost and prediction quality, we follow Saharia et al. [35] by directly conditioning on $\gamma_t$ which allows a flexible choice of diffusion steps and the number of the generation steps is finally set to 100 for testing.

## C. Comparison with the State-of-the-art Methods

In this section, we compare our proposed model with six state-of-the-art (SOTA) radiotherapy dose prediction methods, namely UNet [3], U-ResNet-D [20], DeepLabV3+ [7], C3D [5], PRUNet [6], Mc-GAN [21], and DiffDP [10]. Wherein, C3D utilizes two cascaded 3D UNet as backbone with volumetric data for both input and output, Mc-GAN is a GAN-based model for dose prediction, and DiffDP is the conference version of our method. Quantitative comparison results about the dose scores and APEs of clinic metrics are offered respectively in Table I and Table II, where our proposed DiffDose surpasses the other seven SOTAs with the best overall performance. Concretely, UNet achieves the worst performance, especially 3.751Gy for $|\Delta D_{98}|$ and 0.131Gy for $|\Delta CI|$ on the PTV. With the help of a progressive refinement strategy, PRUNet gains a better performance which reduces the dose score from 2.184Gy to 1.795Gy. Besides, with powerful GAN architecture, Mc-GAN reaches significant performance promotions in several clinical metrics, e.g., 0.0398 for PTV $|\Delta HI|$, 3.573Gy for FHL $|\Delta D_2|$, respectively. DiffDP achieves good performance on PTV, e.g., 1.902Gy for $|\Delta D_{98}|$ and 0.486Gy for $|\Delta D_2|$, but poor performance on OARs. Compared to DiffDP, the proposed method achieves a relatively higher value for PTV $|\Delta D_{50}|$, i.e., 0.479Gy, but it largely reduces FHL $|\Delta D_2|$ by 1.523Gy and FHR $|\Delta D_2|$ by 2.113Gy, respectively. Notably, C3D uses 3D architecture to learn spatial knowledge from 3D volumes and directly predict the 3D dose distribution, thus gaining the second-best dose score. Compared to C3D, the proposed method fed with 2D images still obtains the best results which decreases the dose score by 0.040Gy, PTV $|\Delta D_{mean}|$ by 0.363Gy, and FHL $|\Delta D_{mean}|$ by 1.157Gy, respectively. These results confirm the strong capability of our method in capturing the additional inter-slice information among 2D slices which can serve as a competitive substitute to the 3D ones inside the volumes. Also, comparison of dose errors in Fig. 4 and $|\Delta V_{40}|$ in Fig. 5 demonstrate our proposed method still maintains its superiority with the minimum prediction errors for the OARs.

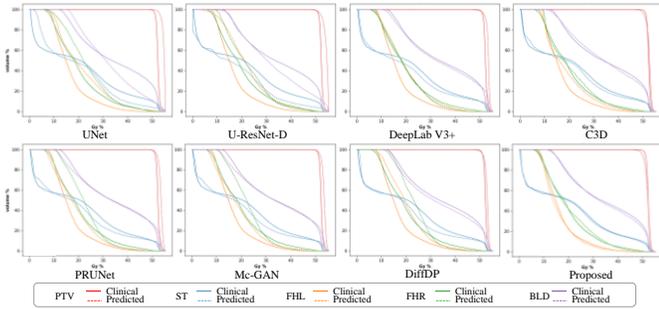

**Fig. 6.** DVH curves for comparison with SOTA methods.

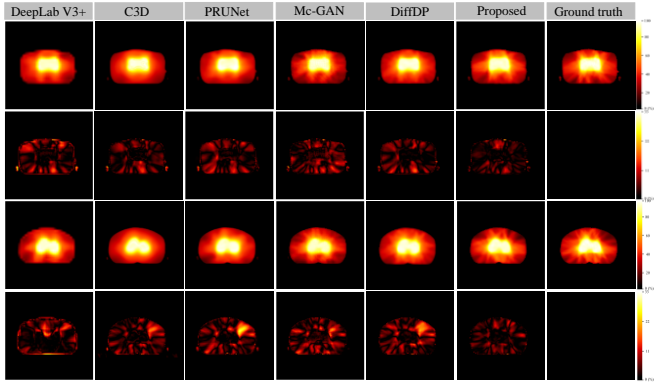

**Fig. 7.** Two groups of visual comparisons with SOTA models. From top to bottom: predicted dose distribution maps and their corresponding difference maps.

In Fig. 6, the DVH curves generated from the compared methods offer a more intuitive representation. As observed, the performance of UNet and U-ResNet-D is relatively unsatisfactory and their disparities to the ground truths are large, especially for the PTV, FHL, and BLD. Although the results of DeepLab V3+ and C3D are more accurate overall, they still suffer from a notable gap with ground truth in terms of FHL. Compared with these methods, the DVH curves of our method best match the ground truth regarding both PTV and OARs, especially for PTV and FHL.

Besides, we present the visual comparison in Fig.7. Notably, the proposed method maintains its leading-edge performance with the darkest difference maps. Notably, the UNet-based methods, i.e., C3D and PRUNet, gain blurred with fewer high-frequency details. Thanks to adversarial learning, Mc-GAN predicts dose maps with more visual texture details, but there are also obvious artifacts that are quite different from the real ones. Differently, the dose maps generated by the proposed method gains the best visual effect with rich high-frequency information, i.e., texture details as well as ray directions.

In summary, our method obtains superior performance in both quantitative metrics and visual qualitative aspects.

*D. Ablation Study*

We conduct incremental ablation experiments to analyze the respective contribution of individual components in our proposed method. Concretely, the experimental arrangements can be summarized as (A) vanilla diffusion model that simply concatenates the anatomical images $x$ and noisy image $y_t$ together as the original input (denoted as Baseline), (B)

TABLE III
QUANTITATIVE RESULTS OF ABLATION MODELS IN TERMS OF DOSE SCORE OF BODY AND $|\Delta CI|$, $|\Delta D_{98}|$, $|\Delta D_{50}|$, AND $\Delta|D_{\text{mean}}|$ OF PTV, SHOWN IN THE MEAN (STANDARD DEVIATION) FORM. THE BEST RESULT OF EACH INDEX IS MARKED IN **BOLD**.

|     | dose score (Gy) | $|\Delta CI|$ | $|\Delta D_{98}|$ (Gy) | $|\Delta D_{50}|$ (Gy) | $\Delta|D_{\text{mean}}|$ (Gy) |
|-----|---|---|---|---|---|
| (A) | 2.080 (0.472) | 0.167 (0.131) | 3.423 (2.165) | 1.794 (0.723) | 1.771 (0.763) |
| (B) | 1.829 (0.257) | 5.42E-2 (5.21E-2) | 2.399 (2.140) | 0.624 (0.384) | 0.628 (0.364) |
| (C) | 1.724 (0.200) | 3.44E-2 (5.74E-2) | 2.059 (2.530) | 0.544 (0.384) | 0.539 (0.427) |
| (D) | **1.615** (0.257) | **3.18E-2** (4.16E-2) | **2.056** (2.537) | **0.479** (0.359) | **0.493** (0.389) |

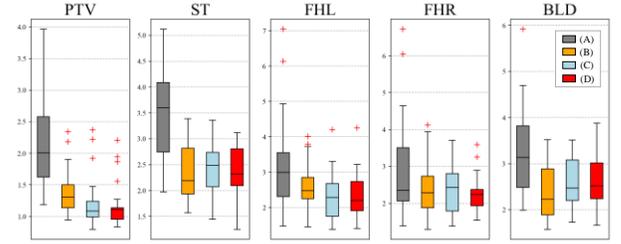

**Fig. 8.** Dose errors of ablation experiments on PTV and OARs.

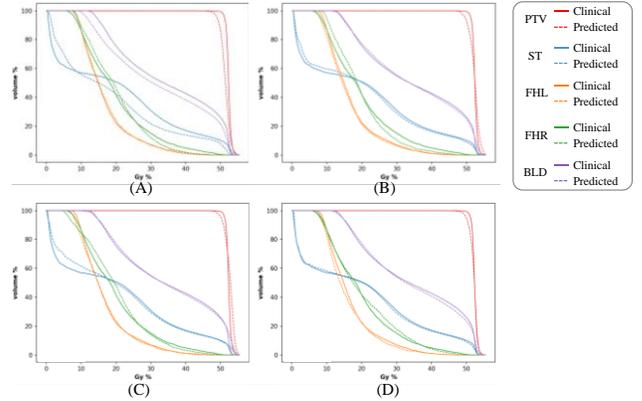

**Fig. 9.** DVH curves for the ablation variant models.

Baseline + structure encoder without I²T block (Baseline + SE w/o I²T), (C) Baseline + structure encoder with I²T block (Baseline + SE w I²T), and (D) Baseline + SE w I²T + adaptive weighted loss (Baseline + SE w I²T + awLoss, Proposed). Table III gives out the quantitative results of PTV while Fig. 8 shows the dose errors of each ablation variant. Besides, we visualize the corresponding DVH curves in Fig. 9.

**Effectiveness of Structure Encoder:** Comparing the results between model (A) and (B) in Table III and Fig. 8, (B) significantly reduces the dose score from 2.080Gy to 1.829Gy, and PTV $|\Delta D_{\text{mean}}|$ from 1.771Gy to 0.628Gy with the help of structure encoder. Correspondingly, for ST, (B) still gains a 1.147Gy decrease in terms of $|\Delta D_2|$. Also, compared to the DVH curve of (A) in Fig. 9, (B) notably reduces the distances to the ground truth, especially for BLD and ST. These results have demonstrated the effectiveness of employing the structure encoder to process the anatomical images rather than a simple concatenation operation.

**Effectiveness of Inter-slice Interaction Transformer Block:** To validate the effectiveness of I²T block, we can simply compare the prediction performance of model (B) and (C). In Table III, (C) further decreases dose score by 0.105Gy and PTV $|\Delta D_{98}|$ by 0.340Gy, respectively. Such enhancements can also be found in the predicted dose distribution of ST, as illustrated in Fig. 9, indicating the effectiveness of the I²T block.

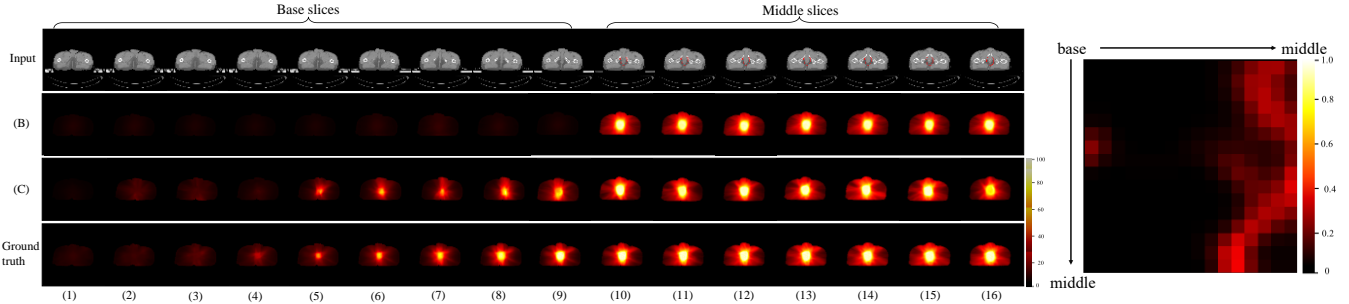

**Fig. 10.** Left: Visual comparisons for model (B) and (C) on different slice parts within a single batch. Input CT images with PTV segmentation contours delineated by red lines, predicted dose distribution maps by (B) and (C), and the corresponding ground truth are shown from top to bottom. Right: Visualization of the attention matrix generated by I²T block where the element $M[j, k]$ evaluates the attention the $j$-th slice pays to the $k$-th slice.

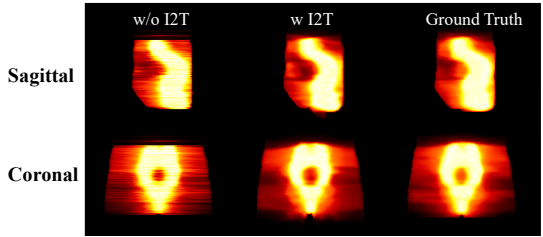

**Fig. 11.** Illustrations of dose maps along the sagittal and coronal direction when the model with or w/o I²T blocks.

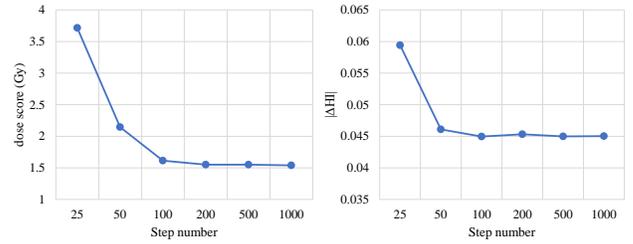

**Fig. 12.** Influence of different numbers of diffusion steps on dose prediction performance in terms of dose score and $|\Delta HI|$.

Besides, to vividly show the capability of I²T block in modeling inter-slice interactions, we display the qualitative results of dose prediction on different slice parts within a single batch in Fig. 10(Left). Input with the continuous images, (B) and (C) achieve quite different prediction performance. Concretely, as for the middle slices with accurate PTV masks (delineated by red lines), i.e., from column (10) to (16), the predictions of both (B) and (C) obtain visually similar results to the ground truth. However, for the base slices which locate close to the PTV boundary but own no PTV structures (i.e., no PTV masks), (B) suffers from notable inconsistency to the ground truth. Differently, equipped with I²T block, (C) gains more consistent predictions in these base slices with the ground truth. These base slices have no PTV segmentation but are still affected inevitably by radiation delivered to nearby PTV area. Such inter-slice information has been effectively captured by the I²T block. Fig. 10(Right) displays the attention matrix generated by the second I²T block of the first residual stage. As seen, the base slices pay high attention to the middle ones. This indicates that while predicting the dose distribution to the base slices with no PTV structures, the I²T block forces the model to pay more attention to the middle slices with clear PTV structures and utilize the inter-slice relationship to predict a more accurate dose distribution for base slices. We also give dose maps along the sagittal and coronal direction in Fig. 11. As seen, with I²T block, the model achieves better perception about inter-slice relationship, thus finally gaining a higher prediction quality.

**Effectiveness of Adaptive Weighted Loss:** To validate the effectiveness of adaptive weighted loss (awLoss), we can compare the results of model (C) and (D). As shown in Table. III, compared to (C), (D) further reduces the dose score to 1.615Gy and $|\Delta D_{50}|$ to 0.479Gy. Fig. 9 illustrates that the DVH curves of (D) closely align with the ground truth mores, showcasing a higher accuracy. This highlights the effectiveness of awLoss in imposing greater penalties for inaccurate predictions pertaining to both PTV and OARs.

**Trade-off between sampling steps and prediction quality:** DiffDose involves multiple steps in the reverse process to generate high-quality predicted dose distribution maps from the Gaussian noise, which is time-consuming. To explore a trade-off between the prediction quality and time cost, we verify the influence of different numbers of reverse steps on the final prediction performance. Concretely, we set the step numbers as 25, 50, 100, 200, 500, and 1000, respectively, and summarize the prediction accuracy in Fig. 12. As seen, the prediction performance is initially enhanced with more generation steps, and then remains stable after 100 steps. So, we set the generation steps to 100 for model testing. With this setting, our model can efficiently predict the dose distribution for each patient with 1.98±0.94min.

## V. CONCLUSION

In this paper, we present DiffDose, a diffusion model-based method for predicting dose distribution maps in radiotherapy planning. DiffDose utilizes a forward process to convert dose distribution maps into pure Gaussian noise in a gradual way and a reverse process to generate the predicted dose maps by progressively removing the noise from pure Gaussian noise.

To incorporate the essential anatomical information, we employ a structure encoder to learn such knowledge inside patients' anatomy images. The structure encoder is further equipped with inter-slice interaction transformer ($I^2T$) blocks, aiming to model long-range dependency information among slices. Furthermore, we introduce an adaptive weighted loss to exert ample attention to PTV and OARs, thus better satisfying their dose requirements. The experiments on the rectum cancer dataset have verified its superiority.